\documentclass{article} 
\usepackage[preprint]{colm2026_conference}

\usepackage{microtype}
\usepackage{hyperref}
\usepackage{url}
\usepackage{booktabs}
\usepackage{graphicx}
\usepackage{cleveref}
\usepackage{siunitx}
\usepackage{tcolorbox}
\usepackage[T1]{fontenc}

\usepackage{lineno}

\definecolor{darkblue}{rgb}{0, 0, 0.5}
\hypersetup{colorlinks=true, citecolor=darkblue, linkcolor=darkblue, urlcolor=darkblue}

\title{When Models Know More Than They Say: Probing\\ Analogical Reasoning in LLMs}

\author{Hope McGovern\textsuperscript{1}\\
\texttt{hem52@cam.ac.uk}\\
 \And Caroline Craig\textsuperscript{2,3}\\
\texttt{cacraig@athenahealth.com}\\ \AND Tom Lippincott\textsuperscript{4}\\ 
\texttt{tom@cs.jhu.edu}\\
\\
\textsuperscript{1}{Cambridge University}\\ \textsuperscript{2}{Northeastern University}\\
\textsuperscript{3}{Athenahealth}\\
\textsuperscript{4}{Johns Hopkins University} \hspace{3cm}
\And Hale Sirin\textsuperscript{4}\\ \texttt{hsirin1@jhu.edu}
}

\begin{document}

\ifcolmsubmission
\linenumbers
\fi

\maketitle

\begin{abstract}
Analogical reasoning is a core cognitive faculty essential for narrative understanding. While LLMs perform well when surface and structural cues align, they struggle in cases where an analogy is not apparent on the surface but requires latent information—suggesting limitations in abstraction and generalisation. In this paper we compare a model’s probed representations with its prompted performance at detecting narrative analogies, revealing an asymmetry: for rhetorical analogies, probing significantly outperforms prompting in open-source models, while for narrative analogies, they achieve a similar (low) performance. This suggests that the relationship between internal representations and prompted behavior is task-dependent and may reflect limitations in how prompting accesses available information.
\end{abstract}

\section{Introduction}\label{introduction}

Humans rely on narratives to explain
causes, encode memory, and convey moral lessons
\citep{BrunerActualmindspossible1986}. 
Despite remarkable advances in language modeling, we still lack benchmarks for assessing whether LLMs acquire the competencies that support narrative understanding, including allusion detection, figurative language production, and
complexity \citep{hamilton2025narrabench}.

\begin{figure}[hb]
    \centering
    \includegraphics[width=3.0in]{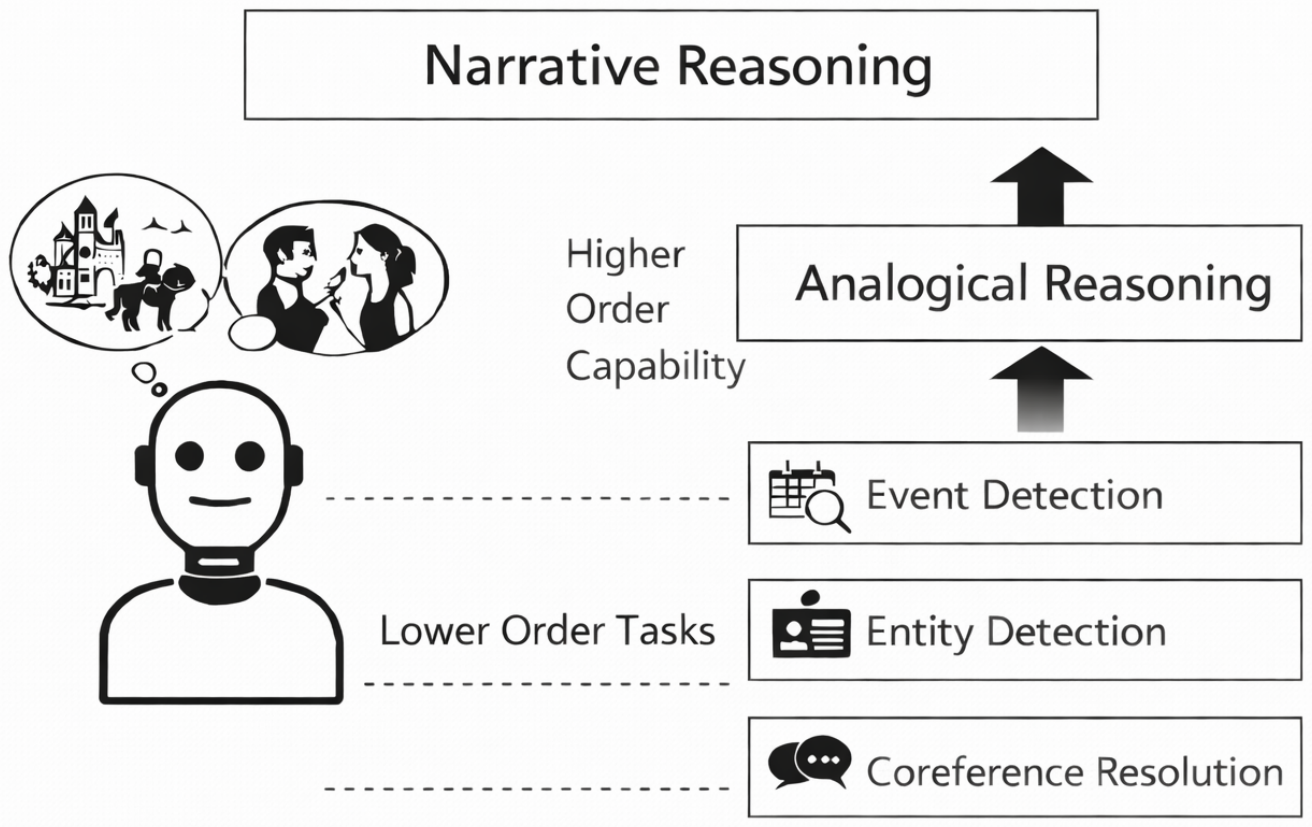}
    \caption{Analogical reasoning is a higher-order capability that requires a combination of lower-level tasks such as entity detection or coreference resolution.}
    \label{fig:illustration}
\end{figure}

Analogical reasoning is a core cognitive faculty of identifying structural similarities between a familiar situation (source) and a new, less understood one (target), allowing knowledge from the source to be mapped and applied to the target \citep{gentneranalogical}. While humans effortlessly perceive
such parallels across diverse episodes, LLMs remain largely untested in this respect.

Recent work highlights narrative understanding as key for model performance \citep{kim2023fantombenchmarkstresstestingmachine, karpinska2024thousandpairsnovelchallenge, hamilton2025narrabench, srivastava2023imitationgamequantifyingextrapolating}.
Pretrained models can encode latent information about entities and relations without
explicit supervision \citep{LiImplicitRepresentationsMeaning2021},
and prompting strategies like chain-of-thought (CoT) have been used as evidence that LLMs can perform
reasoning-like operations. But such results leave open a deeper
question: do LLMs internalize typological structures---the narrative
schemas and rhetorical functions that underlie coherent
storytelling---or are they simply leveraging surface-level correlations
at scale?

Analogical reasoning is defined as the ability to perceive similarities between concepts, situations or events based on (systems of) relations rather than surface phenomena \citep{2012-08871-013, SouratiARNAnalogicalReasoning2024}. Narrative offers a rigorous testbed for analogical reasoning: it
requires integrating causal chains, temporal dependencies, and thematic
abstractions across extended spans of text. Unlike paraphrase or
entailment, recognizing that two passages instantiate the same
\emph{narrative function} despite divergent surface forms reflects a
higher-order cognitive ability. Such abilities are critical not only for
literary analysis and typological exegesis \citep{mcgovern-etal-2025-chiasm}, but also for practical
applications in problem solving, education and scientific discovery \citep{dunbar-scientific}.

If models
truly learn structured representations of text, they should exhibit
efficiencies akin to human narrative understanding: abstraction, reuse
of functional templates, and recognition of rhetorical parallels. If
they do not, this supports the view that despite their scale, LLMs
remain shallow in representation.

\subsection{Our Contribution}\label{our-contribution}

We introduce \textbf{NARB} (\emph{Narrative Analogical Reasoning Benchmark}), a
suite of benchmark tasks designed to probe analogical reasoning in
literary texts.

\begin{itemize}
    \item \textbf{Benchmarking:} We evaluate recent decoder-only LLMs on a hierarchy of tasks from basic narrative role identification to complex rhetorical and structural parallelism.
    \item \textbf{Diagnosis:} We apply interpretability probes to assess which model layers encode narrative and rhetorical information, and to what extent. We compare a model’s probed representations to its prompted performance, showing that for complex narrative analogies, what is achievable with prompting alone is not always decodable from internal states, and vice versa. In this way, our work contributes to ongoing discussions about the interpretability and functional validity of probing, especially for abstract tasks.
    \item \textbf{Findings:} Our results highlight that neither probing nor prompting method alone provides a complete picture of model capabilities, showing that the limitation lies not in the task but in how models translate internal representations into prompted behavior. 
\end{itemize}


\section{Background}\label{background}

\subsection{Diagnostic Probing}
Our method of analysing model internals builds on diagnostic probing,
particularly the \emph{edge probing} framework of
\citet{TenneyWhatyoulearn2019}, which decomposes linguistic tasks into graph edges predicted from hidden representations. They find that pretrained models capture progressively deeper features across layers, from syntax to co-reference.
\citet{TenneyBERTRediscoversClassical2019} further showed a `layerwise progression' in BERT, with syntactic information localised early and semantic features appearing at higher layers.

More recent work complicates this picture.
\citet{HeDecodingProbingRevealing2024} find that grammatical features are distributed throughout GPT-2's layers and vary with sentence complexity.
Critically, \citet{NiuDoesBERTRediscover2022} show that previously reported layer effects may reflect artefacts of position and training dynamics, and \citet{BelinkovProbingClassifiersPromises2022} warns
that information discovered by a probe is not necessarily used by the
model at inference time.

This motivates a core aspect of our analysis: we compare probed representations to \emph{prompted} performance, showing that
for complex analogical tasks, what is decodable from internal states
is not always accessible through prompting. Probing also remains under-explored in literary or narrative contexts, where understanding involves event structure, temporal coherence, and causal reasoning across long-form inputs.

\subsection{Analogical Reasoning for Narrative
Tasks}
The precise mechanisms and
representational structures underlying narrative analogy remain
under-explored, especially in computational settings. \citet{ElsonDetectingStoryAnalogies2012}
introduces a \emph{story intention graph} approach that uses
propositional generalisation over discourse relations. For instance, the
propositions `A Lion watched a fat Bull' and `A Fox observed a Crow' are
abstracted to a shared form like `A predator stalking its prey.'
Their system relies on dependency graphs and logic-based
pattern-matching (via Prolog), incorporating both hypernym
generalisation and temporal sequencing to detect analogies. This method
contrasts bottom-up statistical matching with top-down structural
isomorphisms, offering one of the earliest computational treatments of
analogical narrative structure.


A key insight here is that narrative analogy often involves
category-level parallelism: mapping characters, goals, and events by
type or function rather than surface similarity. 
\citet{SouratiARNAnalogicalReasoning2024} introduce a triplet-based
benchmark dataset, Analogical Reasoning over Narratives (ARN), that
tests whether models can distinguish deep analogies from superficial
resemblance, showing that while LLMs perform well when surface and
structural cues align, they struggle in cases where an analogy is not
apparent on the surface but requires latent information---suggesting
limitations in abstraction and generalisation. 

Outside narrative domains, analogical reasoning has also been probed
through prompting techniques. \citet{WickeUsingAnalogicalReasoning2024}
show that Chain-of-Thought (CoT) explanations for spatial analogies
yield modest improvements in alignment with human judgments. More
compellingly, \citet{YasunagaLargeLanguageModels2024} demonstrate that
\emph{analogical prompting}---asking models to first generate a relevant
exemplar before solving a target problem---outperforms both zero- and
few-shot CoT baselines, particularly in larger models like GPT-4 and
PaLM-2. 

These findings suggest that LLMs \emph{can} sometimes
exhibit analogical reasoning, especially under structured prompting
regimes and at larger scales. However, it remains unclear whether their apparent analogical inferences
reflect genuine conceptual abstraction or sophisticated
pattern-matching, and the absence of explicit structural
priors---such as event schemas or narrative roles---may constrain both the generalisability and interpretability
of these analogies.

\section{Tasks and Datasets}


%



\begin{table}[t]
\begin{center}
\begin{tabular}{lp{0.75\textwidth}}
\toprule
Anchor         &When I remember the challenges I went through when I was starting my business, I break into tears. But I do not regret a thing. I think that the most precious gold goes through the hottest furnace. It made me better. \\
\cmidrule{2-2}
Analogy             &Once upon a time, in a small village, there lived a talented young presenter named Lily. She faced repeated challenges, but each obstacle made her stronger and more resilient, ultimately earning her respect and admiration. \\
\bottomrule
\end{tabular}
\end{center}
\caption{Example narrative parallelism pair from the Analogy dataset}\label{tab:analogy_example}
\end{table}

\begin{table}[t]
\begin{center}
\begin{tabular}{ll}
\toprule
\multicolumn{1}{c}{\bf Latin}  &\multicolumn{1}{c}{\bf English} \\
\textit{satietas sitiret} & \textbf{satiety} might \textbf{thirst}, \\
\textit{uirtus infirmaretur} & \textbf{strength} might be \textbf{weakened}, \\
\textit{sanitas uulneraretur} & \textbf{health} might be \textbf{wounded}, \\
\textit{uita moreretur} & \textbf{life} might \textbf{die}. \\
\bottomrule
\end{tabular}
\end{center}
\caption{Rhetorical Parallelism in Latin: The repeated syntactic and semantic inversion emphasizes paradoxical transformation.}
\label{tab:rhetoric-parallelism}
\end{table}

We consider two reasoning tasks, representing different notions of parallelism:

\textbf{Task 1: Narrative Parallelism.} Whether a model can recognize systematic structural correspondences between complete narratives. Given an anchor story and a set of candidate stories, the task is to identify which candidates are most parallel to the anchor, independent of surface similarity as seen in Table \ref{tab:analogy_example}. Parallel narratives may differ substantially in setting, characters, or vocabulary, but share an underlying schema or functional progression (e.g. temptation–fall–redemption) \citep{mcgovern-etal-2024-detecting-narrative}. This task probes the extent to which models encode high-level narrative structure rather than topical or lexical overlap.

\textbf{Task 2: Rhetorical Parallelism.} Whether a model can recognize localized stylistic and semantic symmetry within a document. Given a span serving as an anchor (e.g., a line from a sermon or poem), the model must rank other spans by their degree of rhetorical parallelism with the anchor. Parallel spans typically instantiate a shared syntactic template or semantic inversion (e.g., paradox or antithesis), but may vary in lexical content as seen in Table \ref{tab:rhetoric-parallelism} (reproduced from \citet{BothwellIntroducingRhetoricalParallelism2023}). Unlike narrative parallelism, this task emphasizes fine-grained form–meaning correspondences over short textual distances.

Appendix \ref{appendix:corpora} describes the corresponding data sets,
Analogical Reasoning over Narratives (ARN) \citep{SouratiARNAnalogicalReasoning2024} and
Augustinian Sermon Parallelism (ASP) \citep{BothwellIntroducingRhetoricalParallelism2023}, in detail along with examples of positive instances from each.



 
\section{Experiments}\label{experiments}

\subsection{Problem Formulation}

We formalize parallelism as an \textbf{anchor-based ranking problem}. While parallelism could be framed as a binary decision -- \textit{are these two spans (or narratives) parallel or not} -- in practice, human judgments of parallelism are comparative: given a reference item, some candidates are \textit{more} parallel than others, even among negatives.

Accordingly, we cast both rhetorical and narrative parallelism as ranking tasks. Each example consists of an anchor $a$ paired with a candidate set $\mathcal{C} = \{c_1, \dots, c_n\}$, partitioned into positives $\mathcal{C}^+$ and negatives $\mathcal{C}^-$ according to gold annotations. A successful model should assign higher scores to true parallel candidates than to non-parallel ones. Evaluation is therefore based on ranking metrics rather than classification accuracy.

Unlike standard triplet setups---which we found trivially solvable in preliminary experiments---our formulation associates each anchor with multiple positives and negatives, and requires ordering the entire candidate pool by degree of parallelism. This preserves within-class variation and supports finer-grained analysis via MRR and MAP.

\subsection{Data Preparation and Sampling}
\textbf{Data cleaning (ARN).} 
We use the ARN dataset described in \cref{appendix-a} and use a filtering method to ensure grammatical acceptability. This reduces the dataset from 1,315 to 872 unique fluent narratives. See \cref{noise-in-user-generated-stories-arn-dataset} for details of the filtering process and the document-level embedding strategy.

\subsubsection{Candidate pool construction}

\textbf{Narrative parallelism.} For each anchor narrative, we construct a candidate pool by sampling a fixed number of narratives that share the same proverb (positives) and narratives that do not (negatives). Positives include both near and far analogies, while negatives include both near and far distractors, ensuring that surface similarity alone is insufficient for high ranking. Unless otherwise stated, we sample $X$ positives and $Y$ negatives per anchor. An example is shown in Table \ref{tab:arn_examples_extremes}.

\textbf{Rhetorical parallelism.}  Each annotated rhetorical set $S = \{s_1, \dots, s_k\}$ gives rise to multiple ranking examples. Each branch $s_i \in S$ serves as an anchor in turn; the remaining branches form the positive candidate set. Negative spans are sampled from the same sermon but outside the annotated parallel set, controlling for topic and discourse context. The average number of spans in a parallel branch is 2.62; for each anchor we sample a substantially larger pool of negatives to produce a non-trivial ranking problem.

\textbf{Data splits.} We partition all datasets into training, validation, and test splits using an 80/10/10 ratio. 
All reported results use 5-fold cross-validation, with metrics averaged across folds and reported as mean $\pm$ standard deviation.

\subsection{Models}

\subsubsection{Embedding Extraction}

We evaluate decoder-only Transformer models from the LLaMA 3 family at three scales: 1B, 3B, and 8B parameters. We select LLaMA 3 due to its strong performance on general capability benchmarks, the availability of multiple model sizes under a shared architecture, and its open-source release via HuggingFace, which enables controlled layer-wise analysis.

For each model, we retain activations from \textit{all} transformer layers. Prior work has shown that linguistic competencies emerge at different depths in Transformer models -- early layers encoding lexical information, intermediate layers capturing syntactic and semantic structure, and later layers reflecting discourse-level properties \citep{TenneyBERTRediscoversClassical2019}. Extending this line of inquiry to decoder-only architectures, we train probes on (i) individual layers, and (ii) a learned scalar mixture over all layers. Comparing these settings allows us to localize where information relevant to parallelism is most strongly represented.

The scalar mixture assigns a learned weight to each layer, producing a weighted sum of layer representations. We contrast the resulting performance with that obtained from probes trained on embeddings extracted from single layers in isolation.

Span representations are obtained via mean pooling over token embeddings within the span. As an ablation, we also evaluate max pooling, finding qualitatively similar trends. Following standard probing practice for decoder-only models, we extract activations from the final token of each span, consistent with evidence that feed-forward layers in Transformers function as key–value memories encoding learned textual patterns \citep{GevaTransformerFeedForwardLayers2021, MengLocatingEditingFactual2023}.

\subsubsection{Scoring Models}

Given an anchor and a candidate span, we evaluate both non-parametric and learned scoring functions to assess rhetorical parallelism. As a strong baseline, we use cosine similarity between span embeddings, testing whether parallelism can be reduced to embedding proximity alone. We additionally train low-capacity learned rankers—a linear model and a shallow MLP—over standard pairwise comparison features derived from the two embeddings. Model capacity is intentionally constrained following best practices in probe design \citep{HewittDesigningInterpretingProbes2019}.

For the rhetorical task, we include distance-based ablations to control for positional confounds, as parallel spans frequently occur near one another in text. These baselines allow us to isolate representational sensitivity to rhetorical structure beyond simple adjacency.

Full mathematical definitions of feature maps, scoring functions, and ablations are provided in \cref{appendix-d}.

\subsection{Evaluation Metrics}

For parallelism tasks, we report standard information retrieval metrics: Mean Reciprocal Rank (MRR), pairwise accuracy, and Mean Average Precision (MAP), which we treat as our primary metric due to its sensitivity to multiple positives. For auxiliary classification tasks (\cref{sec:aux-tasks}), we report F1, AUROC, and accuracy. All metrics range from 0 to 1, with higher values indicating better performance.










\section{Results}\label{results}


\begin{figure}
    \centering
    \includegraphics[width=0.75\linewidth]{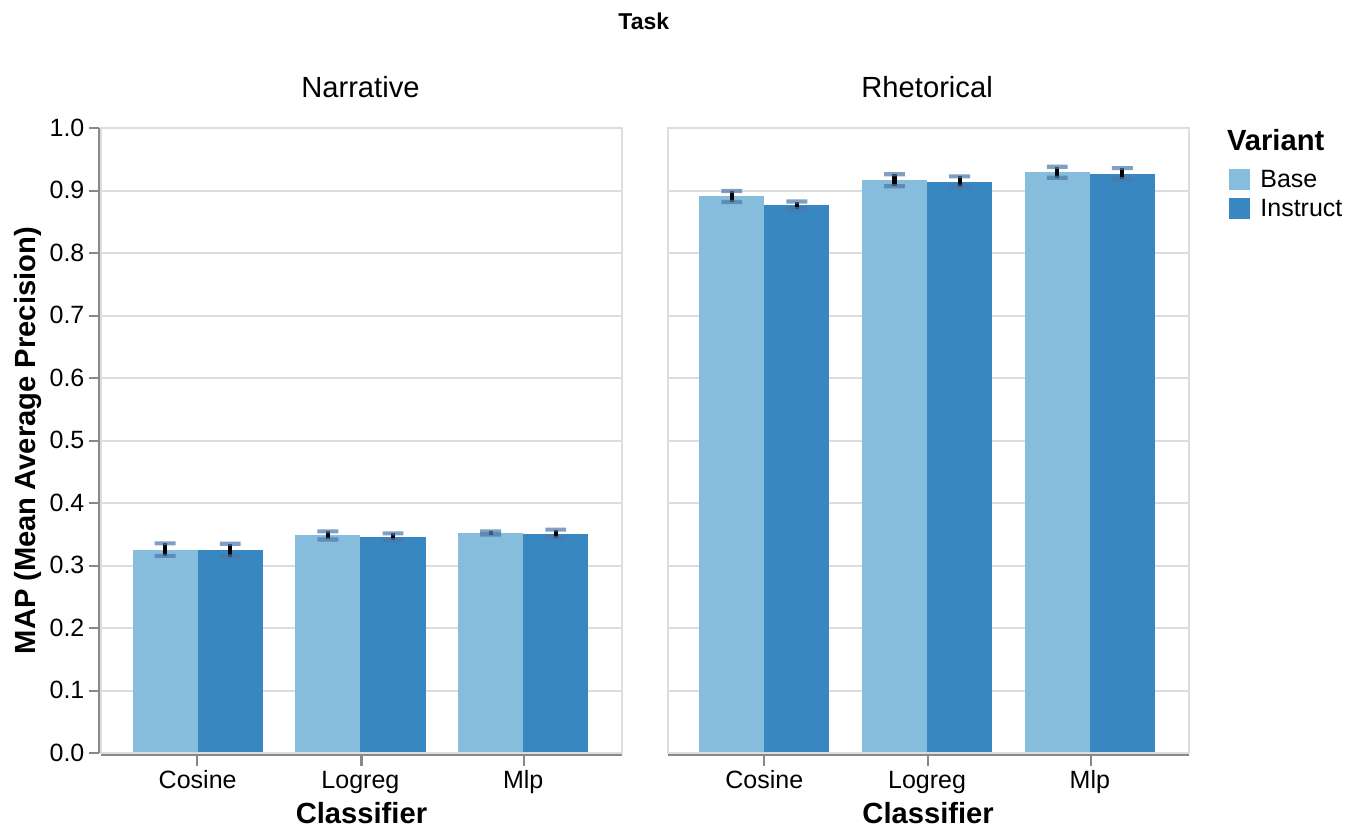}
    \caption{MAP for narrative (left) and rhetorical (right) parallelism tasks across different classifier architectures and model variants on Llama-3.2-1B. Bars show MAP scores (mean ± standard deviation) for three classifier types: cosine similarity (Cosine), logistic regression (Logreg), and multi-layer perceptron (MLP), with separate bars for base and instruction-tuned (Instruct) model variants.}
    \label{fig:map_clf_variants}
\end{figure}


\begin{figure}
    \centering
    \includegraphics[width=.9\linewidth]{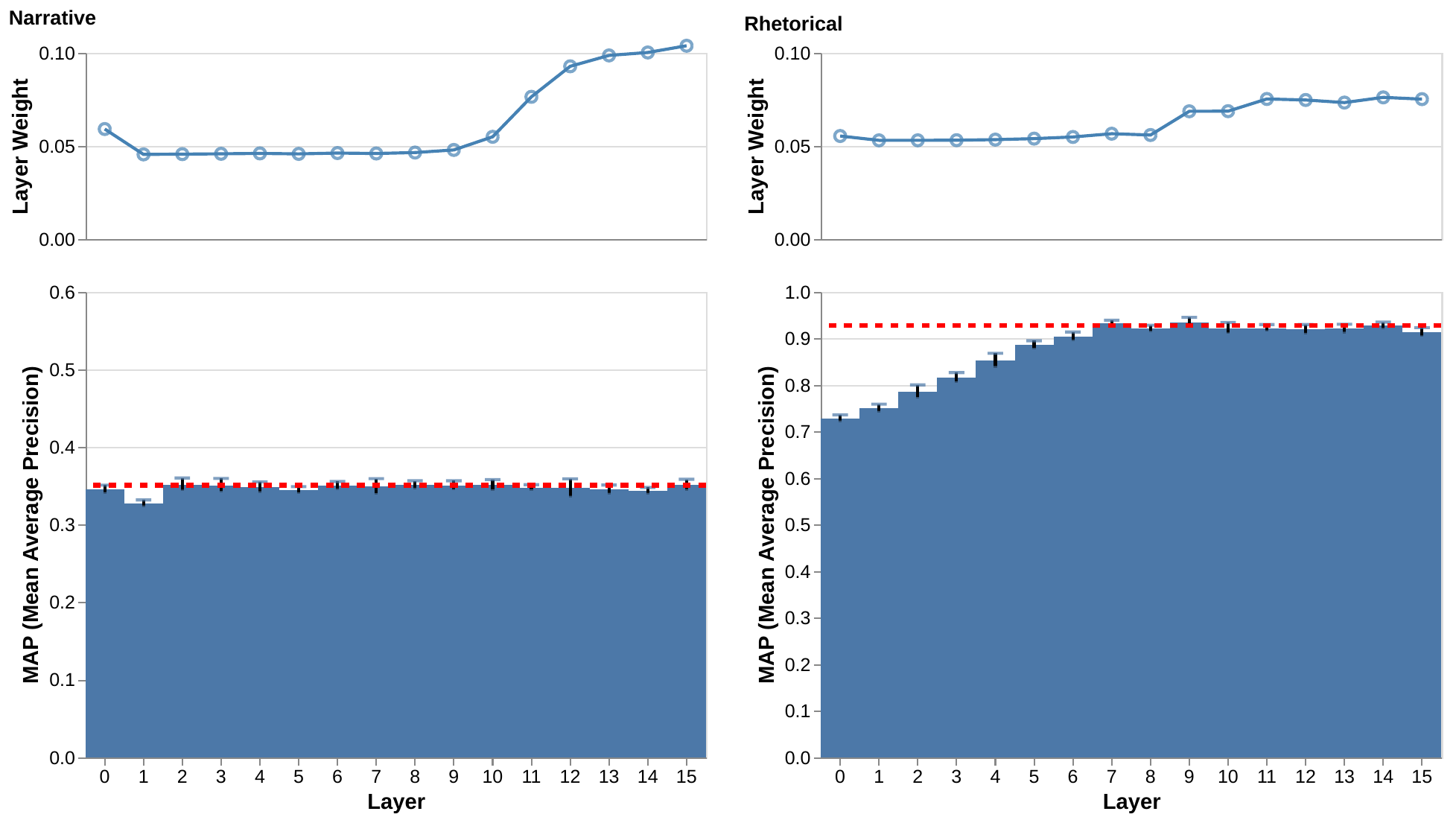}
    \caption{Individual layer performance vs. all-layers configuration on Llama-3.2-1B base model with MLP classifiers. Top panels show learned layer weights from ScalarMix (averaged across cross-validation folds), indicating the relative contribution of each layer when all layers are combined. Bottom panels show mean average precision (MAP) for each individual layer (bars with error bars showing standard deviation) and the all-layers performance (red dashed horizontal line).}
    \label{fig:individual_layers_all_layers}
\end{figure}



\textbf{Narrative parallelism} probing achieves moderate but consistent performance (Figure \ref{fig:map_clf_variants}). The best classifier (MLP) achieves MAP of 0.3506 for base and 0.3493 for instruction-tuned variants, with logistic regression slightly lower (0.3467 and 0.3447) and cosine similarity performing worst (0.3239 and 0.3230). Learned classifiers provide modest improvements over cosine similarity (8--9\% relative improvement), suggesting narrative parallelism benefits from non-linear transformations.

Layerwise analysis reveals that narrative parallelism information is evenly distributed across layers (Figure \ref{fig:individual_layers_all_layers}). Individual layers achieve MAP scores of 0.33--0.35, comparable to the all-layers configuration (0.35), indicating that narrative parallelism does not require integration across multiple layers. This contrasts with rhetorical parallelism, which shows clear progression from early to late layers.


\textbf{Rhetorical parallelism} probing demonstrates exceptional performance, substantially higher than narrative parallelism (Figure \ref{fig:map_clf_variants}). Embedding-based classifiers achieve MAP of 0.93 (MLP) and 0.91 (logistic regression), with cosine similarity achieving 0.89. However, the distance-only baseline achieves MAP of 0.9843 (Table \ref{tab:rhetorical_base_1b_classifiers}), exceeding all embedding-based methods. Combining embeddings with distance (FULL classifier) yields MAP of 0.9845, essentially identical to distance-only, indicating that spatial proximity dominates performance and provides strong evidence for locality dependence in rhetorical parallelism.

Layerwise progression further supports locality dependence (Figure \ref{fig:individual_layers_all_layers}): early layers (0--2) achieve MAP around 0.73--0.75, while later layers (8--15) achieve MAP above 0.90, with peak performance around layers 8--9 (0.93--0.94). This progression suggests that while later layers may capture more abstract patterns, the fundamental locality signal is present even in early layers. The dominance of distance-only performance suggests this signal is primarily structural rather than semantic.


\textbf{Individual layers versus full-model configurations} reveal distinct patterns. For narrative parallelism, individual layers (MAP 0.33--0.35) match all-layers performance (0.35), indicating information is accessible from single layers. For rhetorical parallelism, early layers achieve MAP around 0.73 while later layers exceed 0.90, with all-layers configuration (0.93) performing similarly to the best individual layers.

Learned layer weights from ScalarMix show that narrative parallelism has relatively uniform weights across early layers 
with gradual increases in later layers 
For rhetorical parallelism, weights are lower in early layers 
with a sharp increase around layer 9 
and sustained high weights through layer 15. These patterns align with the performance differences: narrative parallelism relies on distributed semantic representations accessible from multiple layers, while rhetorical parallelism depends primarily on structural locality, with distance-based features (MAP 0.98) dominating over embedding-based features (MAP 0.93).






\section{Auxiliary Tasks}\label{sec:aux-tasks}

Although our primary focus is parallelism, we include a set of simpler auxiliary tasks as sanity checks for our probing framework. These tasks are well-studied, have established annotation standards, and operate over literary text, making them suitable controls for assessing whether our methodology can recover known linguistic and discourse-level distinctions. We consider four auxiliary tasks as an initial benchmark for validating our approach: (1) Event Detection, (2) Entity Detection, (3) Entity Coreference, (4) Quote Attribution. Each task is uniformly cast as a binary classification problem over spans or span pairs, allowing direct comparison across tasks and models. We describe the tasks' datasets, experiments, and results in Appendix \ref{appendix-b}.

\section{Probing vs.\ Prompting}

\subsection{Prompted Ranking Setup}

We compare probing performance to prompted ranking from decoder-only LLMs, including open-source LLaMA3 models (both instruction-tuned and base variants) and closed-source models (GPT-5.2-2025-12-11 and Claude Opus 4.5-20251101). For prompting, we use a fixed set of 20 candidates per example, randomized to avoid spatial cues, and ask models to provide scalar scores between 0.0--10.0 for each candidate along with reasoning for the top-3 highest-scoring candidates. We enforce structured output using Pydantic models. For the rhetorical task, we provide 50 tokens of context and restrict examples to the first branch in each set to ensure the true answer is not included in the context.




\begin{figure}
    \centering
    \includegraphics[width=0.75\linewidth]{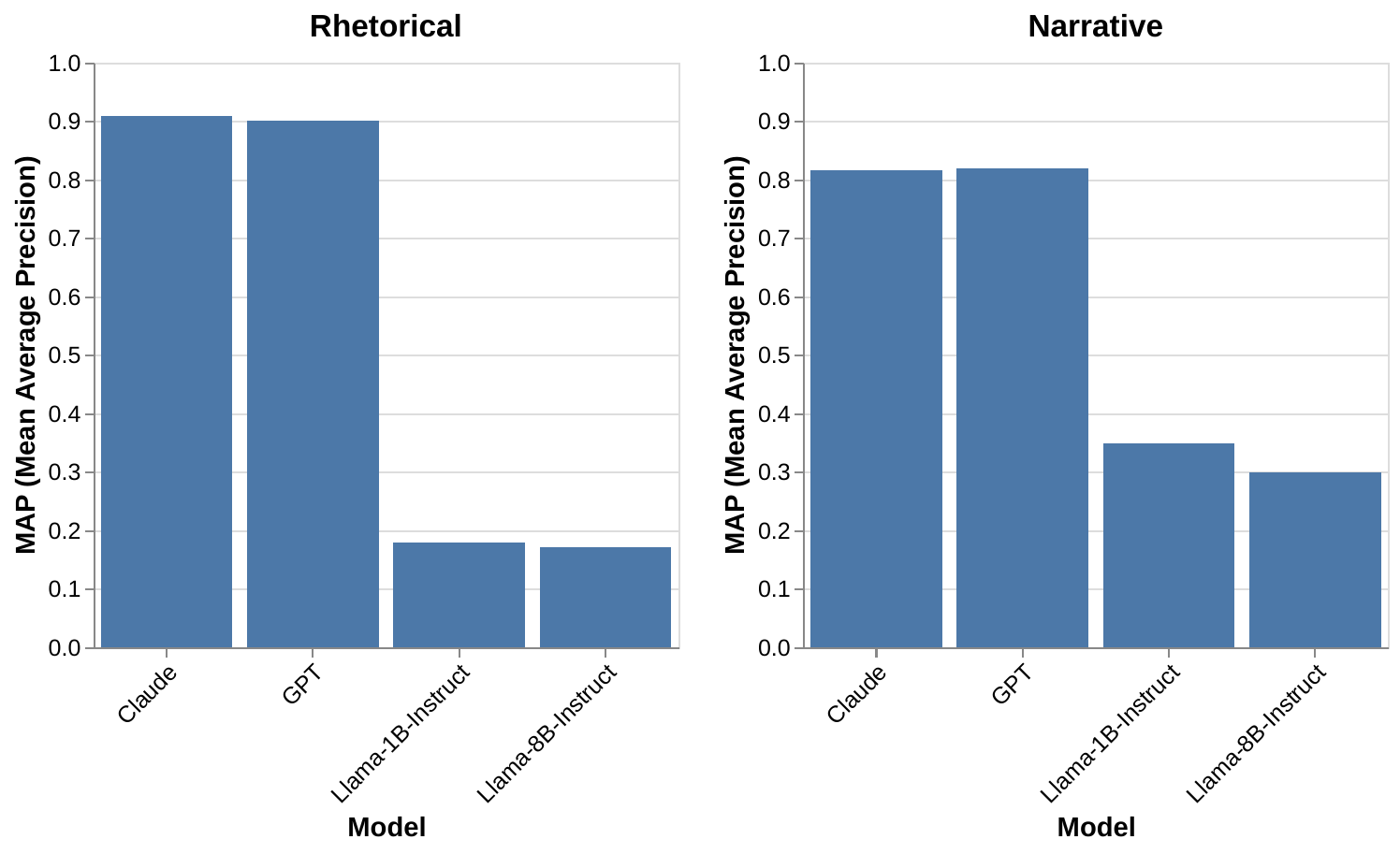}
    \caption{MAP on prompted ranking across different models}
    \label{fig:prompted_ranking}
\end{figure}

\subsection{Comparison Results}

\begin{table}[ht]
\centering
\footnotesize
\begin{tabular}{lcccc}
\toprule
Task & Model & MAP & MRR & Pairwise Acc. \\
\midrule
\textbf{Narrative} & Claude Opus & 0.8158 & 0.9127 & 0.8892 \\
 & GPT-5.2 & 0.8181 & 0.9126 & 0.8921 \\
 & Llama-1B-Instruct & 0.3486 & 0.4487 & 0.4937 \\
 & Llama-8B-Instruct & 0.2987 & 0.3540 & 0.4512 \\
\midrule
\textbf{Rhetorical} & Claude Opus & 0.9084 & 0.9077 & 0.9558 \\
 & GPT-5.2 & 0.9000 & 0.9009 & 0.9753 \\
 & Llama-1B-Instruct & 0.1779 & 0.1921 & 0.4658 \\
 & Llama-8B-Instruct & 0.1710 & 0.1819 & 0.4760 \\
\bottomrule
\end{tabular}
\caption{Prompting experiments on narrative and rhetorical parallelism tasks. Results show that closed-source models (Claude Opus, GPT-5.2) substantially outperform open-source LLaMA models on both tasks, with particularly large gaps on rhetorical parallelism.}
\label{tab:ranking-results}
\end{table}


\begin{table}[h]
\centering
\footnotesize
\begin{tabular}{l S[round-precision=3] @{${}\pm{}$} S[round-precision=2]}
\toprule
Classifier & {MAP (mean $\pm$ std)} \\
\midrule
Cosine & 0.8890 & 0.0088 \\
Distance & 0.9843 & 0.0061 \\
Logreg & 0.9149 & 0.0096 \\
MLP & 0.9278 & 0.0088 \\
Full & 0.9845 & 0.0036 \\
\bottomrule
\end{tabular}
\caption{MAP (mean ± std) for Rhetorical Task, Base Variant, 1B Model}
\label{tab:rhetorical_base_1b_classifiers}
\end{table}

Our comparison reveals task-dependent patterns (Figure \ref{fig:prompted_ranking}). For narrative parallelism, open-source probing and prompting converge (MAP $\approx$0.35), while closed-source models substantially outperform both (GPT-5.2 and Claude Opus: MAP $\approx$0.82). For rhetorical parallelism, the pattern diverges strikingly: open-source models achieve MAP of 0.93 when probed but only 0.17--0.18 when prompted, while closed-source models reach probing-level performance (GPT-5.2: 0.90, Claude Opus: 0.91). The distance-only baseline (MAP 0.98, Table \ref{tab:rhetorical_base_1b_classifiers}) exceeds all other methods, reinforcing the importance of locality for rhetorical parallelism. We discuss the implications of these patterns in \cref{analysis}.


\section{Analysis and Discussion}\label{analysis}

\subsection{Are Probes Relying on Linguistic Information?}
To investigate this question, we applied eight non-LLM-based, linguistic/stylometric methods to the ASP and ARN test sets. These include methods to estimate lexical similarity based on word and N-gram overlaps, syntactic similarity using part of speech (POS) tags, and semantic similarity using sentence embeddings. We find that these methods have varying success in identifying similar and dissimilar documents in the ARN and ASP datasets. We observe better results with the ASP dataset, particularly in recognizing dissimilar documents, though this may be due to the shorter average length of ASP spans, which allow for fewer possible sets of overlaps and POS combinations. Further details and pairplots for both datasets are presented in Appendix \ref{appendix-c}.

\subsection{What Does Probing vs. Prompting Reveal?}

The most striking finding is a task-dependent dissociation between what models \emph{know} (probing) and what they can \emph{do} (prompting). For rhetorical parallelism, LLaMA-3.2-1B-Instruct achieves MAP of 0.93 when probed but only 0.18 when prompted---a fivefold gap---indicating that rhetorical structure is linearly decodable yet inaccessible through instruction-following. That closed-source models achieve probing-level performance (GPT-5.2: 0.90, Claude Opus: 0.91) confirms this is not a task limitation but reflects how open-source models fail to recruit encoded knowledge.

Narrative parallelism presents a different picture: information is both weakly encoded and weakly accessible. Probing and open-source prompting converge at MAP $\approx$0.35, while even closed-source models (MAP 0.82) fall well short of the near-perfect rhetorical scores, suggesting narrative analogy is genuinely harder---requiring abstraction beyond structural patterns.

These patterns have methodological implications. The rhetorical results challenge the assumption that probing reflects usable model capabilities, while the narrative results, where probing and prompting agree, support probing's validity. This task-dependent relationship suggests that probing and prompting should be evaluated together, with their agreement or disagreement providing diagnostic information about how knowledge is encoded and accessed.

\section{Conclusion}

We introduced NARB, a benchmark for analogical reasoning in literary texts, and used it to compare probing and prompting as windows into model capabilities. Our central finding is a task-dependent asymmetry: rhetorical parallelism is strongly encoded (MAP 0.93) yet largely inaccessible via prompting in open-source models (MAP 0.18), whereas narrative parallelism is both weakly encoded and weakly accessible (MAP $\approx$0.35). Neither method alone provides a complete picture---what is decodable from internal states is not always achievable through prompting, and vice versa. These results suggest that evaluation relying solely on prompting may underestimate model capabilities, and that probing and prompting should be used jointly when assessing how knowledge is represented and accessed.






\bibliography{colm2026_conference}
\bibliographystyle{colm2026_conference}

\appendix
\section{Dataset Details}
\label{appendix:corpora}

We evaluate narrative and rhetorical parallelism using two complementary datasets that operationalize parallel structure at markedly different scales.

\begin{figure}
    \centering
    \includegraphics[width=0.6\linewidth]{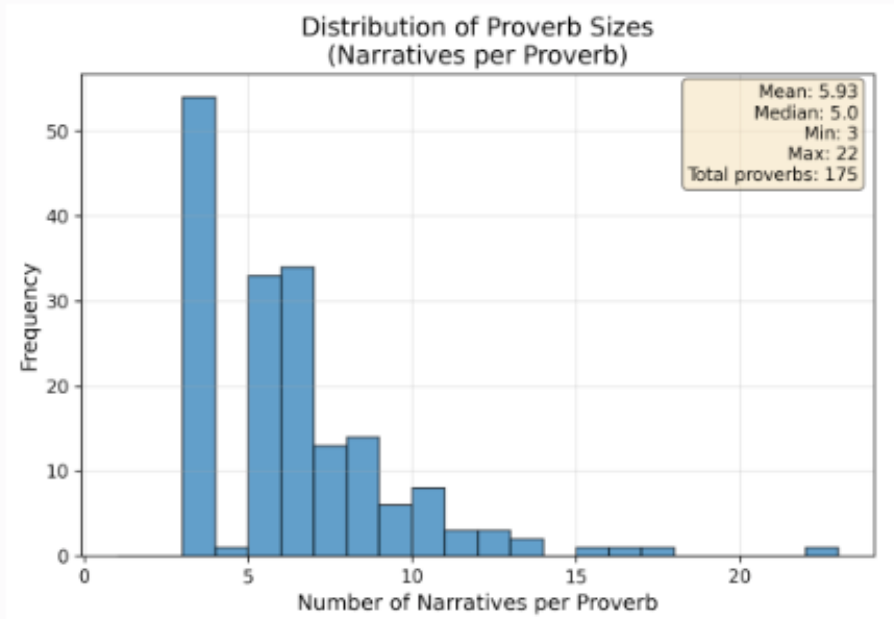}
    \caption{Distribution of proverb sizes based on the number of narratives per proverb}
    \label{fig:proverb_sizes}
\end{figure}

\textbf{Analogical Reasoning over Narratives.} To probe analogical reasoning at the document level, we adopt the Analogical Reasoning over Narratives (ARN) dataset introduced by \citet{SouratiARNAnalogicalReasoning2024}. The underlying narratives are drawn from the ePiC stories dataset \citep{GhoshePiCEmployingProverbs2022}, which contains 2,500 short narratives written by crowdworkers to illustrate a given English proverb (e.g., \textit{Hindsight is 20/20}, \textit{Slow and steady wins the race}). The distribution of proverb sizes in shown in Figure \ref{fig:proverb_sizes}.

\citet{SouratiARNAnalogicalReasoning2024} apply a large language model to extract structured representations from each story, including characters, relations, actions, goals, and locations (collectively termed \textit{surface mappings}), as well as the associated proverb, which functions as a \textit{system mapping}. Using these representations, they construct triplets consisting of an anchor story, an analogous story, and a distractor story. Analogous stories share the same system mapping (i.e., proverb) as the anchor, irrespective of overlap in surface mappings.

The dataset further distinguishes between \textit{near} and \textit{far} cases. A near analogy exhibits substantial overlap in surface features with the anchor (e.g., similar character goals or settings), whereas a far analogy shares only the abstract system mapping. Distractor stories do not share the system mapping: near distractors may resemble the anchor at the surface level but convey a different underlying message, while far distractors differ in both surface features and proverb. The full dataset comprises 1,096 such triplets.

\textbf{Augustinian Sermon Parallelism.} To study rhetorical parallelism at a finer granularity, we use the Augustinian Sermon Parallelism (ASP) dataset introduced by \citet{BothwellIntroducingRhetoricalParallelism2023}. This dataset consists of 80 Latin sermons by Augustine of Hippo, annotated by a domain expert for rhetorical structure.

\begin{figure}
    \centering
    \includegraphics[width=0.6\linewidth]{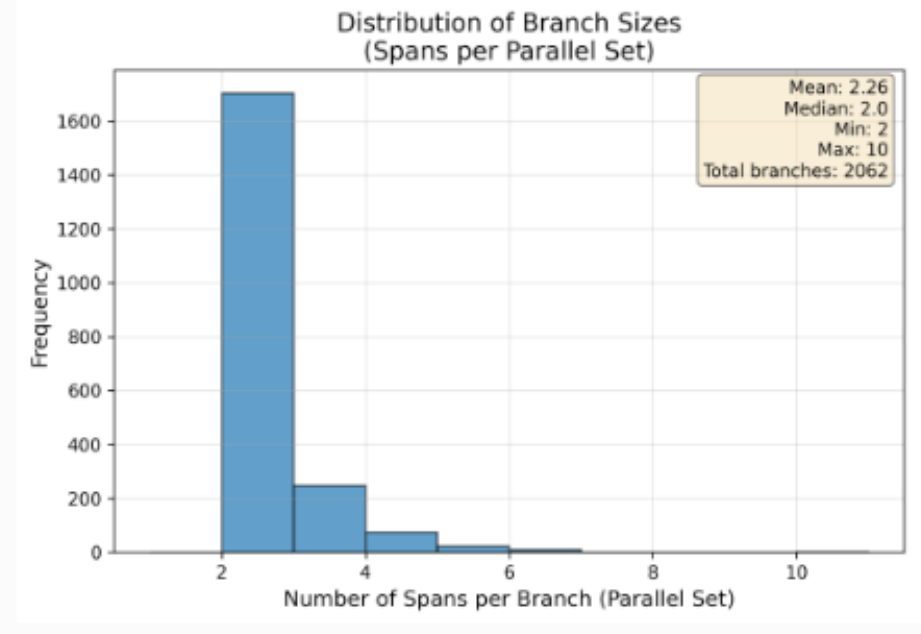}
    \caption{Distribution of branch sizes based on the number of spans per parallel set}
    \label{fig:branch_sizes}
\end{figure}

Annotations identify sets of parallel spans (referred to as \textit{branches}) that jointly instantiate a rhetorical pattern, either synchystic (parallel ordering) or chiastic (inverted ordering). Each set may comprise between two and five spans, often distributed across multiple clauses or sentences. Branch sizes are shown in Figure \ref{fig:branch_sizes} These annotations capture stylistic symmetry at the level of syntax, semantics, and discourse organization, rather than lexical repetition alone.



\section{Additional Result Figures}
\label{appendix:additional_results}

Additional results are shown in Table \ref{tab:classifier_metrics}.



\begin{table}[h]
\centering
\resizebox{\columnwidth}{!}{%
\footnotesize
\begin{tabular}{l l l 
  S[round-precision=3] @{${}\pm{}$} S[round-precision=2]
  S[round-precision=3] @{${}\pm{}$} S[round-precision=2]
  S[round-precision=3] @{${}\pm{}$} S[round-precision=2]}
\toprule
Task & Variant & Classifier
& \multicolumn{2}{c}{MRR}
& \multicolumn{2}{c}{MAP}
& \multicolumn{2}{c}{Accuracy} \\
\cmidrule(lr){1-3}
\cmidrule(lr){4-5}
\cmidrule(lr){6-7}
\cmidrule(lr){8-9}
Narrative & Base & Cosine & 0.4312 & 0.0182 & 0.3239 & 0.0100 & 0.2113 & 0.0051 \\
         &  & Logreg & 0.4537 & 0.0098 & 0.3467 & 0.0065 & 0.2114 & 0.0046 \\
         &  & MLP & 0.4620 & 0.0097 & 0.3506 & 0.0025 & 0.2401 & 0.0040 \\
Narrative & Instruct & Cosine & 0.4302 & 0.0181 & 0.3230 & 0.0102 & 0.2109 & 0.0048 \\
         &  & Logreg & 0.4518 & 0.0104 & 0.3447 & 0.0053 & 0.2111 & 0.0040 \\
         &  & MLP & 0.4560 & 0.0104 & 0.3493 & 0.0066 & 0.2830 & 0.0170 \\
Rhetorical & Base & Cosine & 0.9039 & 0.0099 & 0.8890 & 0.0088 & 0.9625 & 0.0050 \\
         &  & Logreg & 0.9279 & 0.0092 & 0.9149 & 0.0096 & 0.8729 & 0.0082 \\
         &  & MLP & 0.9398 & 0.0080 & 0.9278 & 0.0088 & 0.7088 & 0.0215 \\
Rhetorical & Instruct & Cosine & 0.8933 & 0.0073 & 0.8749 & 0.0060 & 0.9604 & 0.0033 \\
         &  & Logreg & 0.9277 & 0.0079 & 0.9129 & 0.0083 & 0.8787 & 0.0107 \\
         &  & MLP & 0.9382 & 0.0072 & 0.9255 & 0.0088 & 0.6938 & 0.0106 \\
\bottomrule
\end{tabular}
}
\caption{MRR, MAP, and Accuracy (mean ± std) for different classifiers}
\label{tab:classifier_metrics}
\end{table}

\section{Additional Method Details}\label{appendix-a}



\subsection{Noise in User-Generated Stories (ARN dataset)}\label{noise-in-user-generated-stories-arn-dataset}
We observe substantial variation in fluency and grammatical well-formedness across the narratives in ARN, which may introduce noise unrelated to analogical structure. To control for this, we filter narratives using a grammatical acceptability model trained on BliMP-style judgments \citep{KrishnaReformulatingUnsupervisedStyle2020}. We retain only narratives with an acceptability score of at least 0.9, removing ill-formed or degenerate generations (see \cref{appendix-a} for examples). 

\textbf{Document-level embedding strategy.} For both tasks, we embed each entire document -- a sermon or a narrative -- using a decoder-only language model. Span representations are subsequently extracted from these document-level embeddings. This strategy substantially reduces storage and computation costs while preserving contextual information, and ensures that all span embeddings are derived from a consistent global context.



\begin{table*}
\begin{tcolorbox}[width=\linewidth, colback=white, colframe=gray!75!white, title=\textbf{Illustrative Examples from Extremes of the Analogy Dataset}, fonttitle=\bfseries, coltitle=black, boxrule=0.5pt, arc=3pt, left=2pt, right=2pt, top=4pt, bottom=4pt]
\centering
\begin{tabular}{>{\raggedright\bfseries}p{3.5cm} p{\dimexpr\linewidth-3.5cm-4\tabcolsep-2\arrayrulewidth-4pt\relax}}

\toprule
\textbf{Category} & \textbf{Example Content} \\
\midrule
\small
Easiest Analogy 
(Close Analogy, 
Low Distractor Similarity) &

\textbf{Proverb:} \textit{That which does not kill us makes us stronger} \newline

\footnotesize\textbf{Anchor:} When I remember the challenges I went through when I was starting my business, I break into tears. But I do not regret a thing. I think that the most precious gold goes through the hottest furnace. There are great and unforgettable lessons that I learned during that period that I will always cherish. It made me better. \newline

\footnotesize\textbf{Analogy:} Once upon a time, in a small village, there lived a talented young presenter named Lily. She was faced with countless challenges, from technical difficulties and stage fright to harsh criticism and rejection. However, with each obstacle she overcame, Lily grew stronger, honing her skills, building resilience, and gaining the respect and admiration of her audience. Ultimately, her unwavering determination and ability to thrive in the face of adversity transformed her into a renowned and influential figure, inspiring others to embrace their own inner strength. \newline

\footnotesize\textbf{Distractor:} I stained my cedar house this last summer in advance thinking about the consequences of not doing so. If I don’t, the sun fades and cracks the wood, and could let drafts in. \\

\midrule
\small
Hardest Analogy (Far Analogy, High Distractor Similarity) &

\textbf{Proverb:} \textit{Hindsight is always twenty-twenty} \newline

\footnotesize\textbf{Anchor:} Mark was the new CEO of the company. Under adrenaline rush he decided to go after a small startup that he thought would be profitable for the company. However, months later, it was discovered that the startup would not benefit them much but instead it was costing them a fortune to make a bid for the startup. \newline

\footnotesize\textbf{Analogy:} Looking at the past decisions, I put too much oil in the fryer and burned the turkey last year, and the year previous to that and also a couple of months ago. That's also true that at those moments I didn’t know what I’m doing. \newline

\footnotesize\textbf{Distractor:} After becoming the new CEO of the company John decided to change the microchip in the laptop being produced by his company. However he understood that they need to design an entirely new laptop instead of just changing the chip as the new chip won’t be compatible with the old hardware setup. \\
\bottomrule
\end{tabular}
\caption{Examples from the ARN dataset: Easies and hardest analogies}
\label{tab:arn_examples_extremes}
\end{tcolorbox}
\end{table*}

\section{Auxiliary Tasks}\label{appendix-b}

\subsection{Span Classification Framework}

Let a document be represented as a sequence of contextual embeddings

$$E=[e_0,e_1,…,e_n],$$
extracted from a pretrained language model. A **span** is defined as a half-open interval
$$s = [i, j),$$
corresponding to the subsequence $[\mathbf{e}_i, \mathbf{e}_{i+1}, \ldots, \mathbf{e}_{j-1}]$.

Following \cite{TenneyWhatyoulearn2019}, we compute fixed-length span representations by applying a learned linear projection to each $\mathbf{e}_k$ within the span, followed by self-attention pooling across the span window. This yields a single vector representation $h_s$ per span.

Each classification instance consists of either:

1. A single span ($s_1$) (event detection, entity detection) 
2. A pair of spans $(s_1, s_2)$ drawn from a shared discourse context (coreference and quote attribution).

The classifier then predicts a binary label $y \in {0,1}$ indicating whether the span (or span pair) satisfies the task-specific criterion. Importantly, these tasks do \textit{not} require ranking; they serve to verify that the same embeddings and lightweight probes can recover more conventional linguistic distinctions.

\subsection{Dataset and Setup}

We describe the provenance and preprocessing of the datasets used for each auxiliary task. Dataset statistics are summarized in Table \ref{tab:task-stats}.

\begin{table}[ht]
\centering
\small
\resizebox{\linewidth}{!}{%
\begin{tabular}{>{\raggedright}p{0.18\linewidth} >{\centering}p{0.42\linewidth} >{\centering}p{0.16\linewidth} >{\centering\arraybackslash}p{0.16\linewidth}}
\toprule
\textbf{Task} & \textbf{Provenance} & \textbf{\# Unique Docs} & \textbf{Task Type} \\
\midrule
Event Det. & Lit-Bank & 100 documents & Span Classification \\
Entity Det. & Lit-Bank & 100 documents & Span Classification \\
Entity Coref. & Lit-Bank & 100 documents & Span Pair Classification \\
Quote Attr. & Lit-Bank & 100 documents & Span Pair Classification \\
Rhetorical Sym. & Augustinian Sermon Parallelism dataset & 80 sermons & Ranking \\
Narrative Sym. & Analogical Reasoning over Narratives dataset from ePiC stories & 872 narratives & Ranking \\
\bottomrule
\end{tabular}
}%
\caption{\textbf{Dataset description by task.} This table shows the provenance of each dataset alongside the number of unique source documents and task type.}
\label{tab:task-stats}
\end{table}

\subsubsection{Task 1: Event Detection}

We use the event annotations introduced by \cite{SimsLiteraryEventDetection2019}, drawn from the first 2,000 words (210,532 tokens) of 100 literary works in Lit-Bank \citep{Bammanannotateddatasetliterary2019}. The dataset contains 7,849 annotated events. Events include activities, accomplishments, achievements, and changes of state, restricted to asserted (\textit{realis}) events involving a specific entity. Event triggers are single tokens (verbs, adjectives, or nominals).
We extract the annotations and build the positive set of all extracted events. To build the negative set, we randomly sample token spans of length 1 (each event is 1 token) that are not labeled as events from the full document.

\subsubsection{Task 2: Entity Detection}

For entity detection, we use the entity annotations provided by \cite{Bammanannotateddatasetliterary2019}, drawn from the same 210,532 first tokens of the 100 Lit-Bank literary works. The final dataset includes 13,912 entity annotations including people,
natural locations, built facilities, geopolitical entities, organizations and
vehicles.

We extract annotations using the first annotator's labels only, resulting in 11,989 annotations, which represent our positive samples. We randomly sample a set of equal length from the spans that include no entity annotations. The length of each negative sample is randomly chosen between 1 and twice the average length of a positive sample.

\subsubsection{Task 3: Entity Coreference}

For entity coreference, we use \cite{BammanAnnotatedDatasetCoreference2020}, who build on the entity annotations in \cite{Bammanannotateddatasetliterary2019} to annotate coreference mentions of these entities, excluding generic references such as the generic "you". We extract 2,164 coreference mentions and build a positive set by sampling all coreference combinations for the same entity and a negative set by sampling combinations from different entities in the same document.

\subsubsection{Task 4: Quote Attribution}

We derive quote attribution examples from the dataset introduced by \cite{SimsMeasuringInformationPropagation2020}, who leverage the coreference annotations in \cite{BammanAnnotatedDatasetCoreference2020} to attribute 1,765 quotations to their speaker(s), drawing from the same 210,532 first tokens across the Lit-Bank's 100 literary works.

We include all 1,765 in the positive set and build the negative samples by randomly pairing a quotation with a different speaker mentioned in the same document.

\subsubsection{Class Balance and Splits}

Due to the combinatorial nature of span-pair construction, negative examples substantially outnumber positives in all auxiliary tasks. To mitigate extreme class imbalance, we downsample negative examples to match the number of positive examples, using a fixed random seed (seed=42) for reproducibility.

During cross-fold validation (k=5) we use document-level splitting to prevent leakage. Full dataset statistics are reported in Table \ref{tab:task-stats}.

\subsection{Models and Evaluation}

For all auxiliary tasks, we use the same embedding extraction procedure as in the parallelism experiments. Span representations (or concatenated span-pair representations) are passed to either a logistic regression classifier or a shallow MLP. No ranking objective is used.

\begin{figure}
    \centering
    \includegraphics[width=0.6\linewidth]{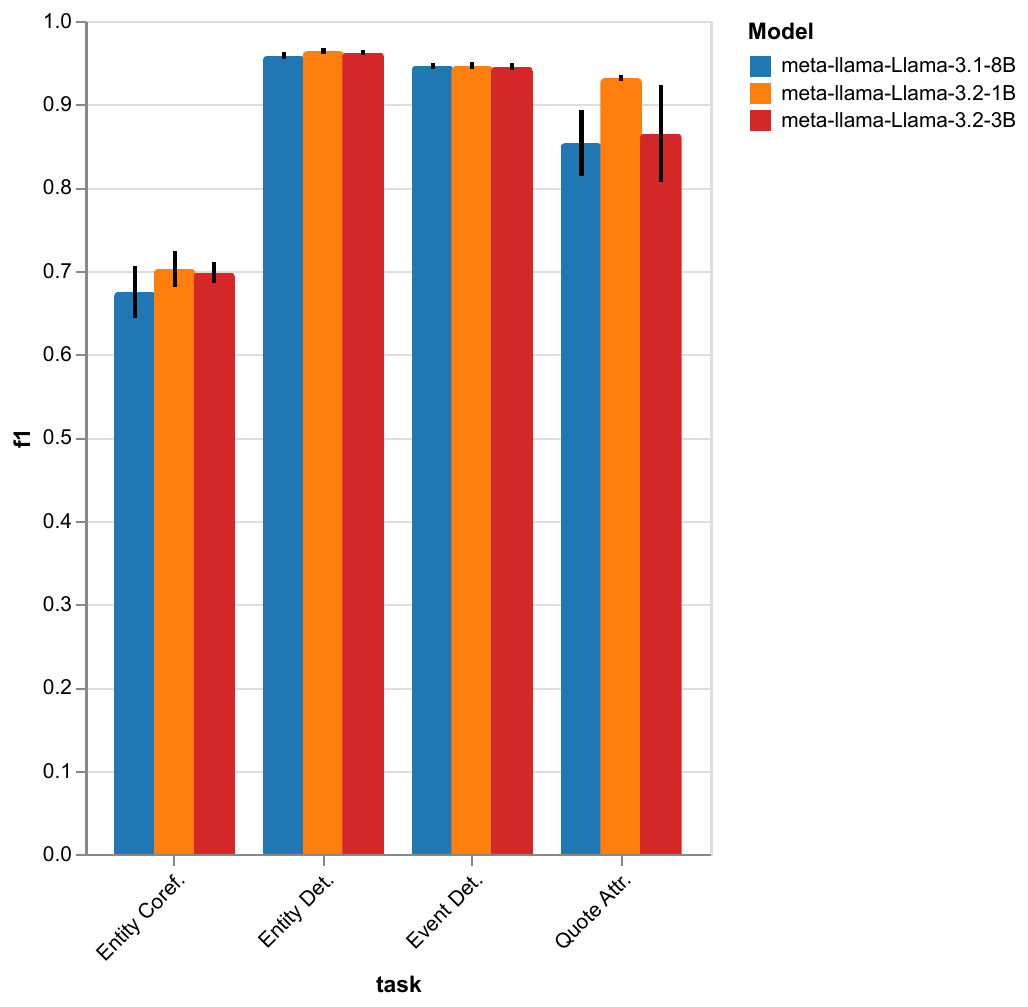}
    \caption{1/3/8B Llama F1 vs Lit-Bank Task (Pooling: Mean, Classifier: MLP)}
    \label{fig:f1_vs_task_baseline_mean_mlp}
\end{figure}

\begin{figure}
    \centering
    \includegraphics[width=0.6\linewidth]{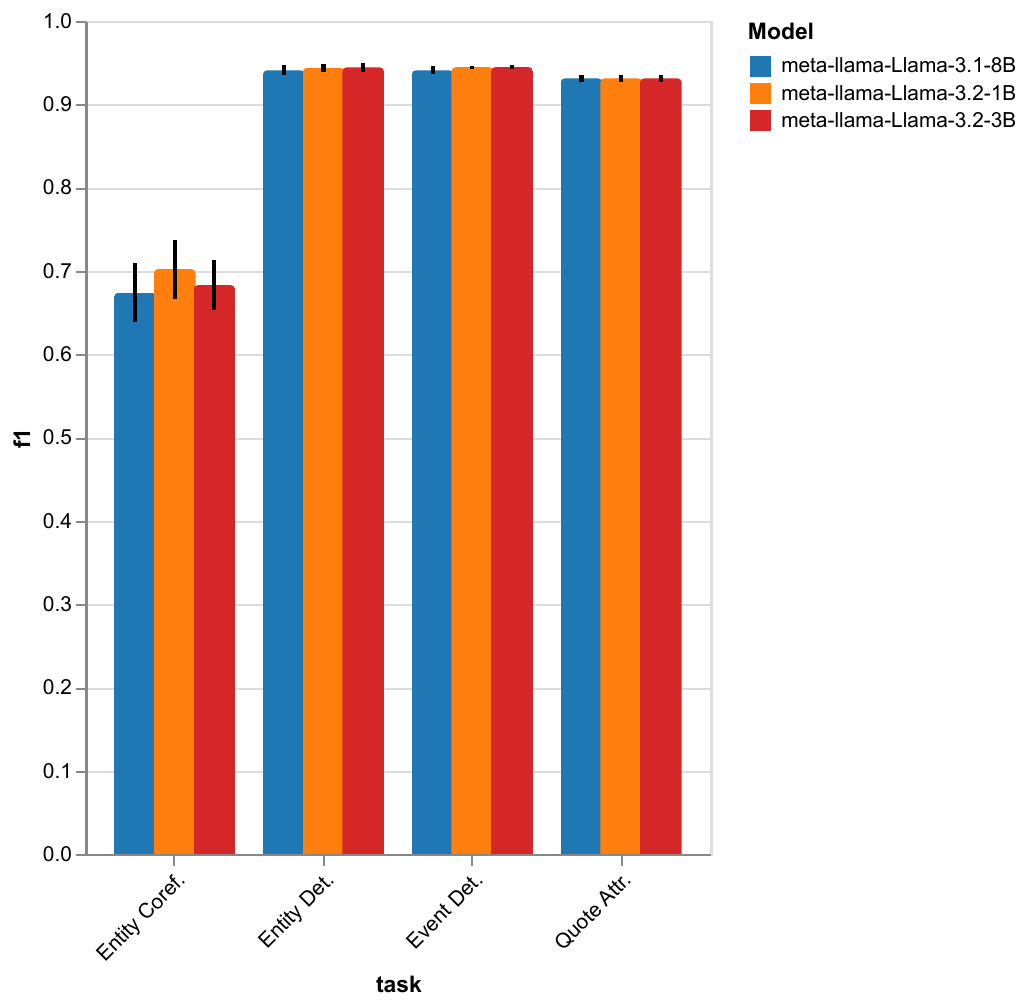}
    \caption{1/3/8B Llama F1 vs Lit-Bank Task (Pooling: Mean, Classifier: LogReg)}
    \label{fig:f1_vs_task_baseline_mean_logreg}
\end{figure}

\begin{figure}
    \centering
    \includegraphics[width=0.6\linewidth]{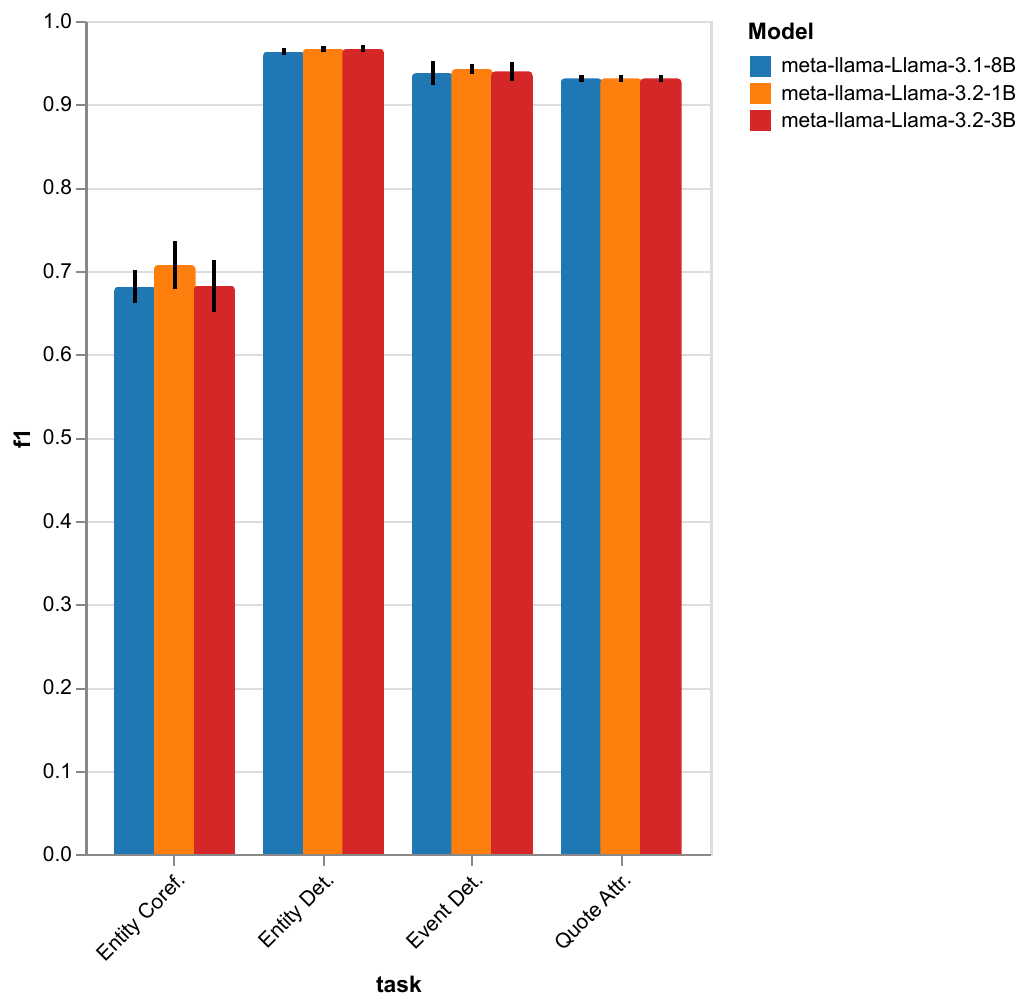}
    \caption{1/3/8B Llama F1 vs Lit-Bank Task (Pooling: Max, Classifier: MLP)}
    \label{fig:f1_vs_task_baseline_max_mlp}
\end{figure}

\begin{figure}
    \centering
    \includegraphics[width=0.6\linewidth]{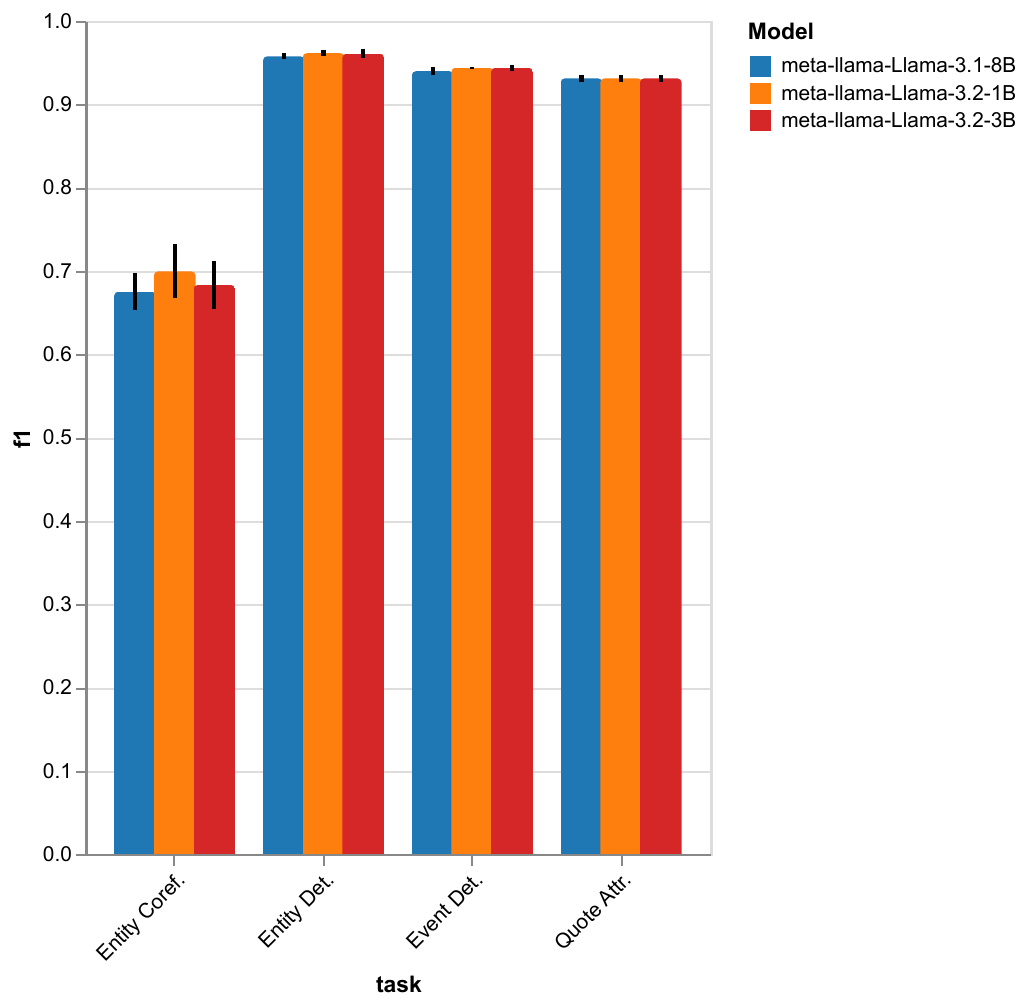}
    \caption{1/3/8B Llama F1 vs Lit-Bank Task (Pooling: Max, Classifier: LogReg)}
    \label{fig:f1_vs_task_baseline_max_logreg}
\end{figure}

We evaluate performance using F1, AUROC, and accuracy, reporting results averaged across cross-validation folds. We show F1 scores in Figures \ref{fig:f1_vs_task_baseline_mean_mlp}, 
\ref{fig:f1_vs_task_baseline_mean_logreg}, \ref{fig:f1_vs_task_baseline_max_mlp}, \ref{fig:f1_vs_task_baseline_max_logreg}.

\subsection{Results}

For the binary classification tasks, we observe best performance on entity detection followed by event detection, quote attribution, and entity coreference (in that order), with a significant gap between the performance of entity coreference and the other three tasks. The smallest model consistently shows the highest performance across all pooling and classification methods, with particular differentiation on quote attribution with mean pooling and MLP classifier. 

\section{Linguistic Analysis}\label{appendix-c}

For all methods, we first normalize the spans (ASP) and documents (ARN) by removing punctuation and lowercasing them. To estimate lexical similarity we calculate the jaccard distance between sets of tokens, lemmatized tokens, and 3-grams, and the BLEU score between sets of tokens. For syntactic similarity we extract POS sequences using stanza and then calculate the edit distance and jaccard distance between pairs of spans and documents. We also build dependency trees using Stanza and compute similarity using networkx's graph edit distance (Latin) and a graph kernel implementation (GraKel) with Weisfeiler Lehman test for graph isomorphism for the longer English documents. Finally, for semantic similarity, we compute the cosine similarity between LaBSE embeddings. 

\begin{figure*}[t]
    \centering
    \includegraphics[width=\textwidth]{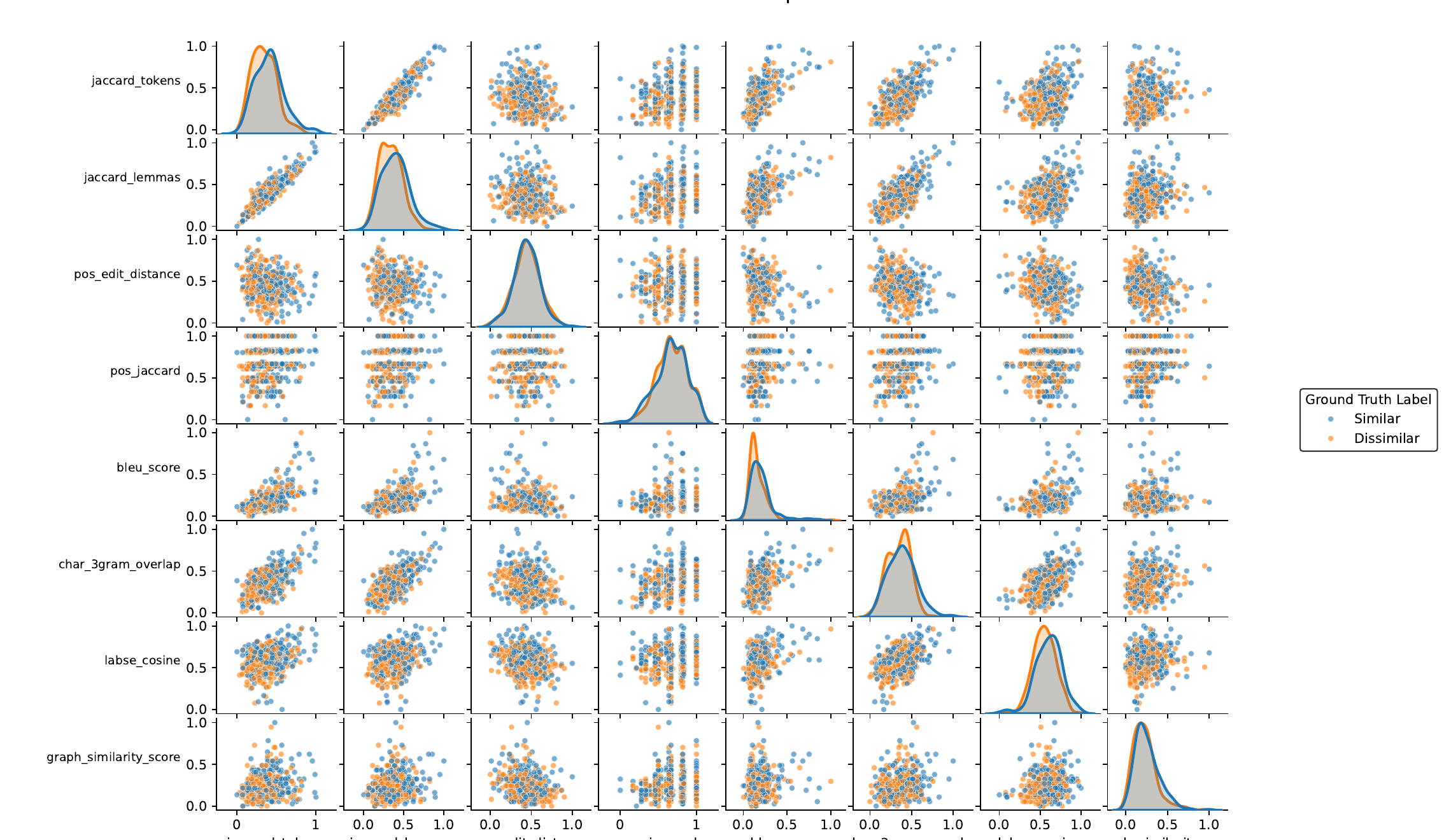}
    \caption{Similarity Scores across 446 ARN Document Pairs (Normalized with Min-Max Scaling) Using Non-LLM-Based Methods}
    \label{fig:arn_pairplots}
\end{figure*}

\begin{figure*}[t]
    \centering
    \includegraphics[width=\textwidth]{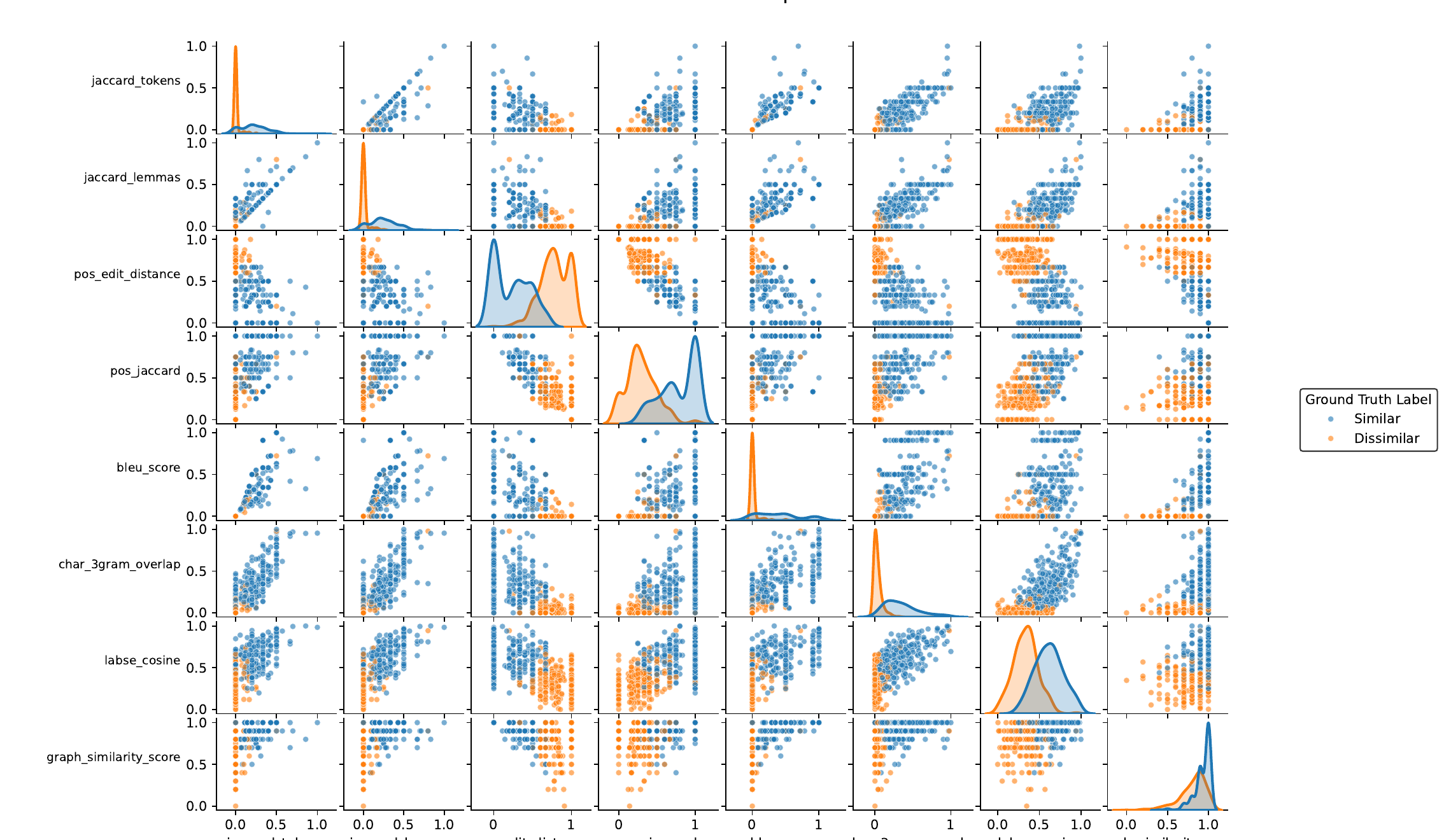}
    \caption{Similarity Scores across 564 ASP Span Pairs (Normalized with Min-Max Scaling) Using Non-LLM-Based Methods}
    \label{fig:asp_pairplots}
\end{figure*}

In the ARN dataset we find similar distributions of scores across both positive and negative ground truth labels, with a tendency toward either a right-skewed or a normal distribution (figure \ref{fig:arn_pairplots}). In the ASP dataset we observe differentiation in the distributions of scores by label. Dissimilar spans have lower similarity scores across all metrics except the edit distance on dependency trees. Similar spans likewise have higher similarity scores except in the four lexical similarity metrics (jaccard distance on tokens, lemmas, 3-grams and BLEU score), where they show a more uniform distribution (Figure \ref{fig:asp_pairplots}).

\section{Scoring Models and Training Details}\label{appendix-d}

\subsection{Span Embeddings}
Let $h_a, h_c \in \mathbb{R}^d$ denote the embeddings of an anchor span $a$ and a candidate span $c$, respectively. Span embeddings are obtained via mean pooling over token-level embeddings within each span. As an ablation, we also evaluate max pooling and observe qualitatively similar trends. For decoder-only models, we extract activations from the final token of each span, following standard probing practice and prior evidence that Transformer feed-forward layers encode salient textual patterns \citep{GevaTransformerFeedForwardLayers2021, MengLocatingEditingFactual2023}.

\subsection{Feature Representations}
We define a standard pairwise feature map for comparing spans:
\[
\phi(a,c) = [h_a;\, h_c;\, |h_a - h_c|;\, h_a \odot h_c] \in \mathbb{R}^{4d},
\]
where $\odot$ denotes element-wise multiplication.

\subsection{Scoring Functions}

\paragraph{Cosine similarity baseline.}
As a non-parametric baseline, we compute cosine similarity between span embeddings:
\[
s_{\text{cos}}(a,c) = \cos(h_a, h_c).
\]

\paragraph{Learned embedding rankers.}
We consider two low-capacity learned scorers over the pairwise feature representation. The first is a linear model,
\[
s_{\text{lr}}(a,c) = w^\top \phi(a,c),
\]
and the second is a shallow multilayer perceptron,
\[
s_{\text{mlp}}(a,c) = \mathrm{MLP}(\phi(a,c)).
\]
Model capacity is intentionally constrained following best practices in probe design \citep{HewittDesigningInterpretingProbes2019}.

\paragraph{Distance-based ablations (rhetorical task only).}
To control for positional confounds, we introduce distance-based baselines. Let $\Delta(a,c)$ denote the token distance between spans. The distance-only representation is defined as
\[
x_{\text{dist}}(a,c) = [\Delta(a,c),\, |\Delta(a,c)|].
\]
We additionally evaluate a combined embedding--distance representation,
\[
\phi_{\text{full}}(a,c) = [\phi(a,c);\, \Delta(a,c);\, |\Delta(a,c)|],
\]
which tests whether embeddings capture rhetorical structure beyond adjacency.

\subsection{Training Objective}
All learned scorers are trained using a pairwise ranking loss. Given an anchor $a$, a positive candidate $p$, and a negative candidate $n$, the objective is
\[
L = -\log \sigma\big(s(a,p) - s(a,n)\big),
\]
which encourages positive candidates to be assigned higher scores than negatives. We employ in-batch negatives throughout training and optimize all models using Adam with standard hyperparameters.

\end{document}